%% file: D-GAN_CVPR.tex
\documentclass[10pt,twocolumn,letterpaper]{article}

\usepackage{cvpr}
\usepackage{times}
\usepackage{epsfig}
\usepackage{graphicx}
\usepackage{amsmath}
\usepackage{amssymb}
\usepackage{float}
\usepackage{pdfpages}

\pdfoutput=1
\usepackage[pagebackref=true,breaklinks=true,letterpaper=true,colorlinks,bookmarks=false]{hyperref}

\cvprfinalcopy 

\def\cvprPaperID{3152} 

\begin{document}
	
	\title{Differential Generative Adversarial Networks: Synthesizing Non-linear Facial Variations with Limited Number of Training Data}
	
	\author{Geonmo Gu, Seong Tae Kim, Kihyun Kim, Wissam J. Baddar, and Yong Man Ro
		\\
		Image and Video Systems Lab., School of Electrical Engineering, KAIST, South Korea\\
		{\tt\small \{geonm, stkim4978, noddown, wisam.baddar, ymro\}@kaist.ac.kr}}

\maketitle

\begin{abstract}
In face-related applications with a public available dataset, synthesizing non-linear facial variations (e.g., facial expression, head-pose, illumination) through a generative model is helpful in addressing the lack of training data. In reality, however, there is insufficient data to even train the generative model for face synthesis. In this paper, we propose Differential Generative Adversarial Networks (D-GAN) that can perform photo-realistic face synthesis even when training data is small. Two discriminators are devised to ensure the generator to approximate a face manifold, which can express face changes as we want. Experimental results demonstrate that the proposed method is robust to the amount of training data and synthesized images are useful to improve the performance of a face expression classifier.
\end{abstract}

\section{Introduction}
The recent success of deep learning has been coupled with the abundance of data availability and big data. It has been shown that deep learning is well generalized when the training data contains a large amount of variations  \cite{krizhevsky2012imagenet, taigman2014deepface, schroff2015facenet, simonyan2014very}. In many cases, collecting large scale datasets suitable for training deep neural networks is very challenging and costly. In some cases, data labeling requires to be performed by experts in a specific domain (e.g., doctors are required to annotate medical images). In order to mitigate the challenge of data collection, a lot of research efforts have been devoted to developing data augmentation methods (e.g., over-sampling \cite{krizhevsky2012imagenet}, horizontal flipping \cite{yang2015mirror}, rotating \cite{xie2015holistically}, photometric transformations \cite{simonyan2014very}).\\
\indent In face related applications (such as face recognition and facial expression recognition), although the aforementioned methods helped to improve the performance of deep neural networks \cite{xu2014integrate, kim2017deep, mollahosseini2016going, kim2017multi}, still the lack of many non-linear variations (e.g., identity, head-pose, illumination, and expression variations) in training dataset could not be addressed via conventional augmentation methods.
\begin{figure}[t]
	\centering
	\setlength{\abovecaptionskip}{3pt}
	\setlength{\belowcaptionskip}{-12pt}
	\includegraphics[width=1.0\hsize]{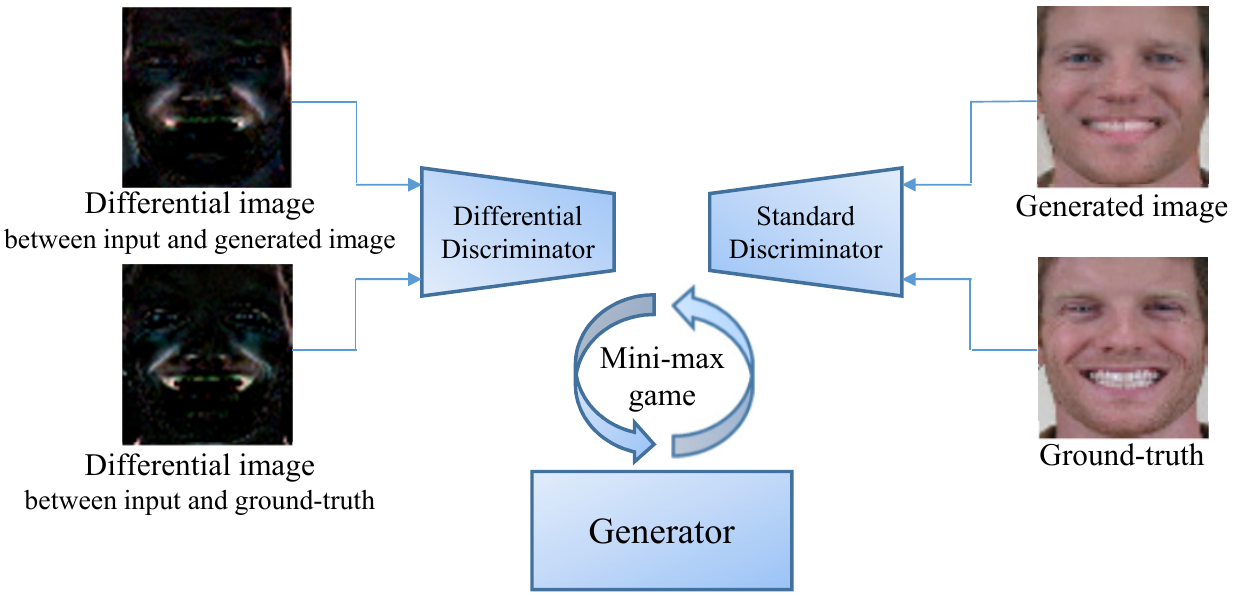}
	\caption{Our GAN-based model for facial expression synthesis. The generator tries to mimic the facial change between input image (e.g., neutral expression) and ground-truth (e.g., happiness) and generate photo-realistic images through two mini-max games with two discriminators.}
	\label{fig1}
	\vspace{-0.4cm}
\end{figure}

\indent To address non-linear augmentation for facial variations, image synthesis methods have been proposed to generate a large number of variations from a given face image. The recent advance of generative adversarial networks (GANs) \cite{goodfellow2014generative} has influenced researches to investigate face image synthesis as photo-realistic images \cite{radford2015unsupervised,zhao2016energy,arjovsky2017wasserstein,berthelot2017began,huang2017beyond,zhang2017age}. The methods of \cite{radford2015unsupervised,zhao2016energy,arjovsky2017wasserstein,berthelot2017began} generated random face images with noise signal as input. The generated random face images could not be controlled so that the method were not suitable to augment non-linear variations of faces. A TP-GAN \cite{huang2017beyond} improved the accuracy of face recognition by synthesizing the frontal face with rotated face as input. Since the TP-GAN was designed to perform many-to-one image synthesis, it could not be applied to many-to-many or one-to-many image synthesis. The authors in \cite{zhang2017age} proposed a conditional adversarial autoencoder (CAAE) to approximate a face manifold to achieve face age progression and regression. The CAAE could be applied to synthesis of other non-linear variations (e.g., expression, head-pose, and illumination). However, we observed that the CAAE could not approximate the face manifold with a small amount of training dataset.\\
\indent In this paper, we propose a new GAN-based model, which is named as differential generative adversarial networks (D-GAN). The proposed model enables the approximation of desired face manifolds. If a face manifold is well trained, we could find the point corresponding to the identity of the input face on the manifold and obtain smoothly changed images traversing on the manifold along with the direction of face changing from that corresponding point. To achieve this goal, we propose a new differential discriminator, which takes a differential image between input and generated image as input of discriminator. In cooperating with standard discriminator \cite{goodfellow2014generative}, the differential discriminator lets a generator try to mimic the facial change of the differential image in a manner of mini-max game so that the generator could effectively synthesize a photo-realistic image.\\
\indent Since the differential image contains only the changed information while the identity is removed, the differential discriminator can concentrate only on the change between the input and generated image. Therefore, as described in Figure \ref{fig1}, the standard discriminator regularizes the generator to synthesize the photo-realistic image and the differential discriminator regularizes the generator to find the right direction of face changing on the face manifold. We have observed that the proposed method is powerful to approximate the face manifold compared to the case of using only the standard discriminator especially under very small dataset. In addition, our method does not need any sophisticated training strategy \cite{larochelle2009exploring,pan2010survey,ding2017exprgan,shrivastava2016learning} to train the generative model even with a small amount of training dataset.\\
\indent The contribution of the proposed method can be summarized as the followings:
\begin{enumerate}
	\itemsep-0.2em 
	\item We propose the novel network architecture in order to approximate the face manifold for non-linear augmentation.
	\item We show that our method can synthesize smoothly changed image and control the specific parts of face (e.g., lips and eyes) through the proposed D-GAN.
	\item In terms of training strategy, our proposed method can be trained with single end-to-end training even with a small amount of training dataset.
	\item We demonstrate that non-linear augmented data can improve the performance of a classifier in terms of accuracy.
\end{enumerate}

\section{Related work}
\noindent\textbf{Generative adversarial networks} Goodfellow et al. \cite{goodfellow2014generative} proposed a novel framework for estimating generative models through an adversarial learning process, which is referred as GAN. Using the random vector $\textbf{z}$  sampled from distribution $p_\textbf{z}$, the generator $G$ produces a fake sample $G(\textbf{z})$. The discriminator $D$ tries to distinguish between real samples $\textbf{y}$ from distribution $p_\textbf{y}$  of ground-truth and fake sample $G(\textbf{z})$. The generator and discriminator are trained by optimizing and adversarial mini-max game. The generator is trained to produce a real sample through the following equation:
\begin{align}
\begin{split}
\underset{G}{\text{min}}\text{ }\underset{D}{\text{max}}\text{ }\mathcal{ L}(G,D)={}&\mathbb{E}_{\textbf{y} \sim p_\textbf{y}}[\log (D(\textbf{y}))]+\\
&\mathbb{E}_{\textbf{z} \sim p_\textbf{z}}[\log (1-D(G(\textbf{z})))].
\label{eq:1}
\end{split}
\end{align}
\indent Recently, many GAN-related studies including face-related applications  \cite{radford2015unsupervised,zhao2016energy,arjovsky2017wasserstein, berthelot2017began,huang2017beyond, tran2017disentangled,zhang2017age,ding2017exprgan,antipov2017face}, published and showed impressive results. DCGAN \cite{radford2015unsupervised}, EBGAN \cite{zhao2016energy},  WGAN \cite{arjovsky2017wasserstein}, and BEGAN \cite{berthelot2017began} generated random face images with noise signal as input. TP-GAN \cite{huang2017beyond} and DR-GAN \cite{tran2017disentangled} improved the performance of face recognition by synthesizing the frontal face with rotated face as input. The authors in \cite{ding2017exprgan} proposed ExprGAN to control expression intensity. The authors in \cite{antipov2017face} proposed Age-cGAN that changes input image to a desired age. 

\noindent\textbf{Image synthesis using traversing the manifold} Image synthesis through traversing the manifold showed impressive results \cite{yeh2017semantic,Ledig_2017_CVPR,zhu2016generative,upchurch2016deep,Shu_2017_CVPR,zhang2017age}. In \cite{yeh2017semantic} and \cite{Ledig_2017_CVPR}, GAN-based models were proposed to learn a traversing from one manifold to another for inpaining and super-resolution, respectively.
A simple interpolation method in the manifold could be effective to image synthesis \cite{zhu2016generative,upchurch2016deep}. However, this method is limited in practical use for non-linear augmentation. For example, Zhu et al. \cite{zhu2016generative} used an user constraint for texture synthesis and image manipulation. It could be difficult to express non-linear variations, such as facial expression and head-pose, as the user constraint used in \cite{zhu2016generative}. Upchurch et al. \cite{upchurch2016deep} used sample images similar to input image. If sample images are not sufficient, face editing on various subject is difficult while maintaining the identity of input image. The authors in \cite{Shu_2017_CVPR} presented face editing by traversing the manifold of latent spaces. However, blurry artifacts were found in synthesized face images. The authors in \cite{zhang2017age} proposed CAAE for face aging, which approximated the face manifold in the latent space and could be applied for other task. We observed CAAE could not approximate the face manifold with a small amount of training dataset. Compared to previous works, our proposed method can produce photo-realistic synthesized face images for non-linear augmentation and approximate the face manifold, even with a small amount of training data.

\section{Differential generative adversarial networks}

\subsection{Network architectures}
\begin{figure*}[!t]
	\small
	\begin{center}
		\includegraphics[scale=0.54] {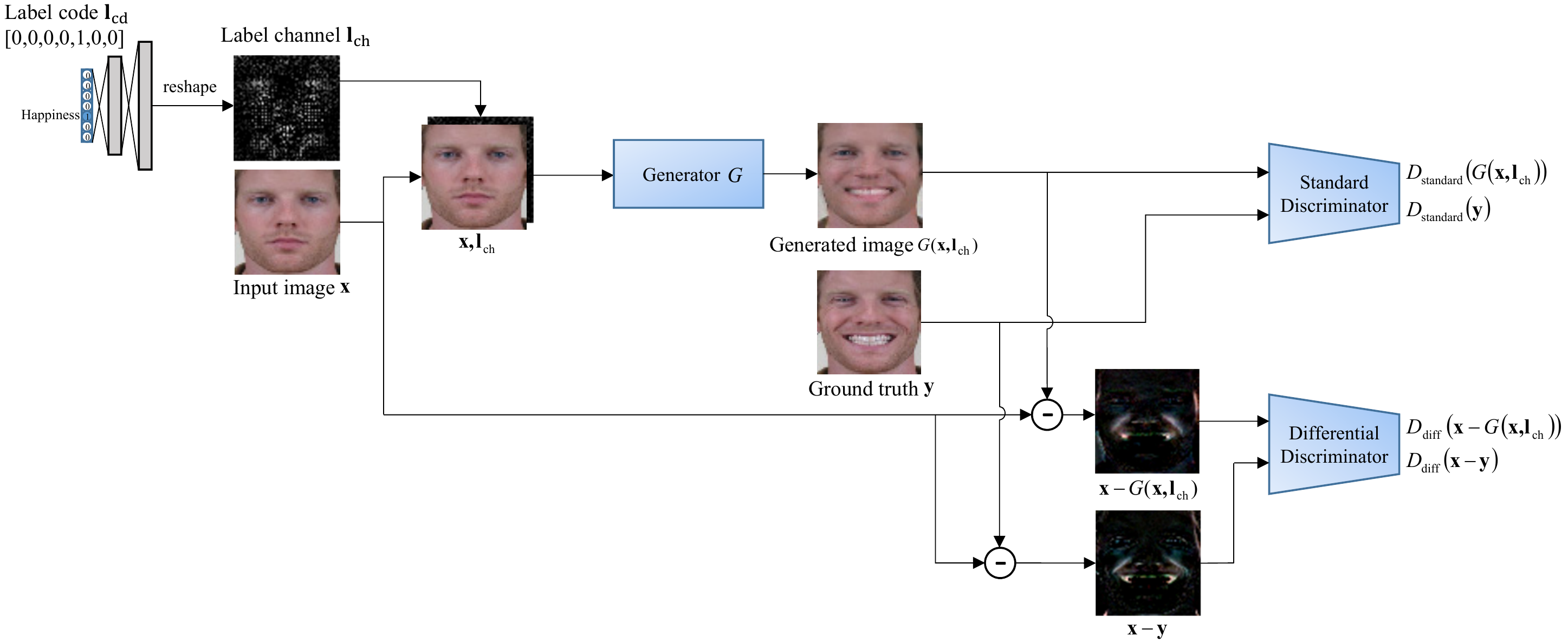}
		\vspace{-0.4cm}
	\end{center}
	\caption{Overall network architecture of D-GAN.}
	\label{fig2}
	\vspace{-0.4cm}
\end{figure*}
\begin{figure}[!ht]
	\centering
	\setlength{\abovecaptionskip}{3pt}
	\setlength{\belowcaptionskip}{-12pt}
	\includegraphics[width=1.0\hsize]{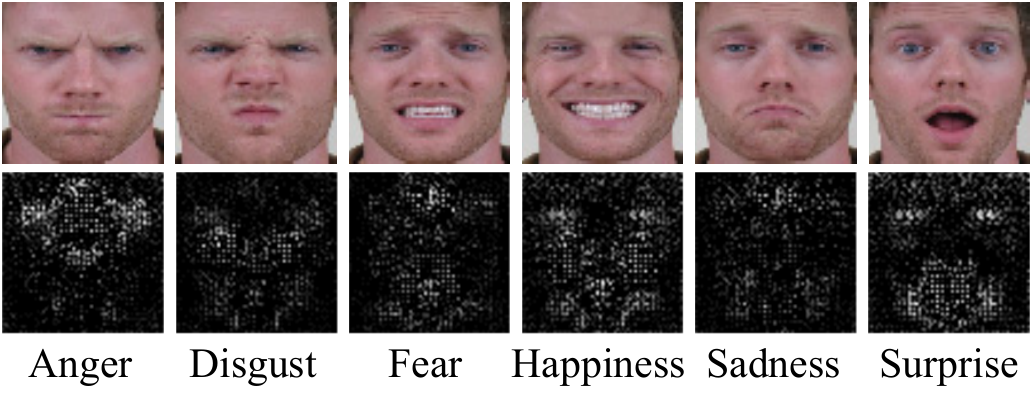}
	\caption{The upper row is ground-truth images and the bottom row is label channels. Note that label channels indicate the spatial position where the face needs to be changed to express the target facial expression.}\label{fig3}
	\vspace{-0.5cm}
\end{figure}
\noindent\textbf{Label channel} The overall network architecture is described in Figure \ref{fig2}. In order to make the generated target image, our proposed model takes an input image $\textbf{x} \in \mathbb{R}^{64 \times 64 \times 3}$ and a label code $\textbf{l}_{\text{cd}} \in \mathbb{R}^N$, where $N$ is the number of labels. The label code $\textbf{l}_{\text{cd}} \in \mathbb{R}^N$ is passed through two fully connected layers with leaky ReLU \cite{maas2013rectifier}. Through the first fully connected layer, the label code   is embedded to a 256-dimensional vector. After that, the 256- dimensional vector is embedded to a 4096-dimensional vector through the second fully connected layer. The 4096-dimensional vector is transformed into a  spatial map $\textbf{l}_{\text{ch}} \in \mathbb{R}^{64\times64\times1}$, which is called as ‘label channel’. The input image $\textbf{x}$ and the label channel $\textbf{l}_{\text{ch}}$ are concatenated and induced to a generator as input. The label channel $\textbf{l}_{\text{ch}}$ is not a simple image reshaped from 4096-dimensional vector. Rather, it is supposed to be trained to indicate the spatial position where the face needs to be changed in an unsupervised manner. Figure \ref{fig3} shows examples of label channels for six facial expressions. In this paper, we can synthesize partial face changes through the combination of label channels. As such, a combined facial expression (e.g., lips are smiled while eyes are angered) can be synthesized.

\noindent\textbf{Generator and discriminator} A U-Net structure \cite{ronneberger2015u} is used as a generator to connect low level layers in encoder and high level layers in decoder by skip connection. U-Net structure allows maximizing the utilization of low-level information to express the details of face change. As previous works \cite{isola2016image,zhu2017unpaired}, noise is provided in the form of the dropout \cite{srivastava2014dropout} on the first two layers in the decoder of the generator with 0.5 rate. Beside the generator, there are two discriminators, which are structurally same but have different functions. One is the standard discriminator $D_{\text{standard}}$ \cite{goodfellow2014generative} to force the generator $G$ to produce photo-realistic images and the other is the differential discriminator $D_{\text{diff}}$, which we propose in this paper. The differential discriminator regularizes the generator to take into account the facial change between input $\textbf{x}$ and ground-truth $\textbf{y}$, which provide complementary effect to achieve successfully synthesized target face images. In detail, we follow architecture guidelines in \cite{radford2015unsupervised,isola2016image}. Both number of convolution layers in encoder part and decoder part of the generator are six. The number of convolution layers in the discriminator is five. BatchNorm \cite{ioffe2015batch} is not applied to the first layer in decoder and discriminator.

\subsection{Discriminators for approximating face manifolds}
Our generative model approximates the face manifold by two mini-max games with two discriminators (the standard discriminator and the differential discriminator). The standard discriminator takes a generated image $G(\textbf{x},\textbf{l}_{\text{ch}})$ or a ground-truth image $\textbf{y} \in \mathbb{R}^{64\times64\times3}$ and performs a mini-max game with following objective functions:
\begin{align}
\mathcal{L}_{D_{\text{standard}}}={}&-\mathbb{E}_{\textbf{y} \sim p_\textbf{y}}[\log(D_{\text{standard}}(\textbf{y}))] \label{eq:2} \\
&-\mathbb{E}_{\textbf{x} \sim p_\textbf{x}}[\log(1-D_{\text{standard}}(G(\textbf{x},\textbf{l}_\text{ch})))], \nonumber \\
\mathcal{L}_{G_{\text{standard}}}={}&-\mathbb{E}_{\textbf{x} \sim p_\textbf{x}}[\log(D_{\text{standard}}(G(\textbf{x},\textbf{l}_\text{ch})))]. \label{eq:3}
\end{align}
\noindent The standard discriminator tries to minimize $\mathcal{L}_{D_{\text{standard}}}$ to distinguish between the ground-truth image and the generated image. The generator tries to minimize $\mathcal{L}_{G_{\text{standard}}}$ to mimic the data distribution $p_\textbf{y}$ of the ground-truth image $\textbf{y}$. Since $\mathcal{L}_{D_{\text{standard}}}$ and $\mathcal{L}_{G_{\text{standard}}}$ are formulated without penalizing the difference between input and output, the generator only cares that the generated image looks like the ground-truth image $\textbf{y}$.

\indent In order to add a constraint that penalizes the difference between the input $\textbf{x}$ and the generated image $G(\textbf{x},\textbf{l}_{\text{ch}})$, we devise the differential discriminator. The differential discriminator is structurally the same with the standard discriminator but it takes a differential image such as $\textbf{x}-\textbf{y}$ and $\textbf{x}-G(\textbf{x},\textbf{l}_\text{ch})$. The mini-max game between the differential discriminator and the generator is performed through the following objective functions:
\begin{align}
\mathcal{L}_{D_{\text{diff}}}={}&-\mathbb{E}_{\textbf{x} \sim p_\textbf{x},\textbf{y} \sim p_\textbf{y}}[\log(D_{\text{diff}}(\textbf{x}-\textbf{y}))] \label{eq:4} \\
&-\mathbb{E}_{\textbf{x} \sim p_\textbf{x}}[\log(1-D_{\text{diff}}(\textbf{x}-G(\textbf{x},\textbf{l}_\text{ch})))], \nonumber \\
\mathcal{L}_{G_{\text{diff}}}={}&-\mathbb{E}_{\textbf{x} \sim p_\textbf{x}}[\log(D_{\text{diff}}(\textbf{x}-G(\textbf{x},\textbf{l}_\text{ch})))]. \label{eq:5}
\end{align}
\noindent The differential discriminator tries to minimize $\mathcal{L}_{D_{\text{diff}}}$ to distinguish between $\textbf{x}-\textbf{y}$ and $\textbf{x}-G(\textbf{x},\textbf{l}_\text{ch})$. The generator tries to minimize $\mathcal{L}_{G_{\text{diff}}}$ to mimic the data distribution of ground-truth differential data $\textbf{x}-\textbf{y}$. Unlike $\mathcal{L}_{G_{\text{standard}}}$ in Eq. \ref{eq:3}, $\mathcal{L}_{G_{\text{diff}}}$ forces the generator to concentrate only on the facial change between the input $\textbf{x}$ and the generated image $G(\textbf{x},\textbf{l}_{\text{ch}})$. Therefore, the differential discriminator can provide complementary effect with the standard discriminator. As a result, the differential discriminator, which focuses on the facial change, allows the generator to approximate the face manifold effectively by taking into account the facial change. In Section \ref{sec4.3}, we demonstrate that our two discriminators and the generator are effective in approximating the face manifold even with a small size of training data.

\subsection{Training D-GAN with total objective functions}
As many previous studies \cite{isola2016image,pathak2016context}, we employ a reconstruction loss such as L1 distance in addition to $\mathcal{L}_{G_{\text{standard}}}$ and $\mathcal{L}_{G_{\text{diff}}}$.
\begin{align}
\mathcal{L}_{\text{Recon}}=\mathbb{E}_{\textbf{x} \sim p_\textbf{x}}[||\textbf{y}-G(\textbf{x}, \textbf{l}_\text{ch})||_1]. \label{eq:6} 
\end{align}
\noindent The mixture of adversarial losses and $\mathcal{L}_{\text{Recon}}$ allows the generator to get closer to the ground-truth in view of pixel-level as well as to fool two discriminators.\\
\indent Overall, the generative model approximates the face manifold by using two adversarial losses and reconstruction loss. Total objective functions are as follows:
\begin{align}
\mathcal{L}_{D_{\text{total}}}&= \mathcal{L}_{D_\text{diff}} + \mathcal{L}_{D_\text{standard}}, \label{eq:7} \\
\mathcal{L}_{G_{\text{total}}}=\lambda_\text{diff} \mathcal{L}_{G_\text{diff}} +&\lambda_\text{standard} \mathcal{L}_{G_\text{standard}}+\lambda_\text{Recon} \mathcal{L}_{\text{Recon}},\label{eq:8}
\end{align}
\noindent where $\lambda_\text{diff}$, $\lambda_\text{standard}$, and $\lambda_\text{Recon}$ are hyper-parameters to control each loss term. As a result, during one iteration of training process, the two discriminators are updated to minimize $\mathcal{L}_{D_{\text{total}}}$ and then the generator is updated to minimize $\mathcal{L}_{G_{\text{total}}}$.

\section{Experiments}

\begin{figure}[t]
	\centering
	\setlength{\abovecaptionskip}{3pt}
	\setlength{\belowcaptionskip}{-12pt}
	\includegraphics[width=1.0\hsize]{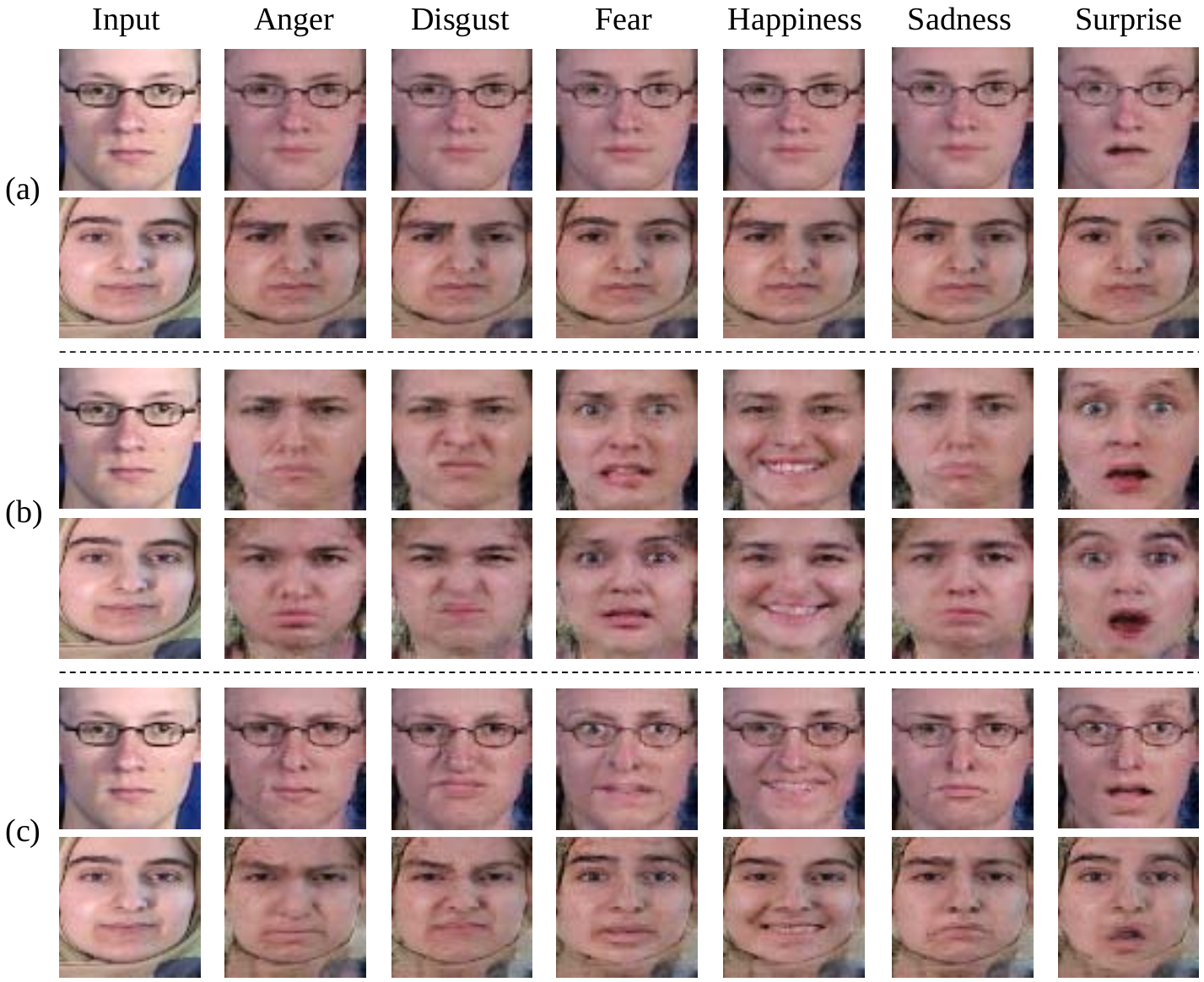}
	\caption{Generated six facial expression images with only standard discriminator (a), only differential discriminator (b), and both standard discriminator and differential discriminator (c).}\label{fig4}
	\vspace{-0.5cm}
\end{figure}
\subsection{Datasets}
To verify the effectiveness of the proposed method, we performed comprehensive experiments. We employed public datasets to train and test our generative model. Datasets used in the experiments are as follows:

\noindent \textbf{PNAS dataset:} This dataset consists of 230 subjects with 22 expressions (the 21 compound expressions plus neutral expression) \cite{du2014compound}. This dataset is captured under supervised by Action Unit (AU) defined by Ekman and Friesen \cite{ekman1977facial}. Out of the 22 expressions, in our experiments, we used the six basic expressions (e.g., anger, disgust, fear, happiness, sadness, surprise) and neutral expression. The total images of those selected subset were 1,610 images. 1,610 images were used as training dataset for our generative model for facial expression synthesis.

\noindent \textbf{MMI dataset:} This dataset consists of 30 subjects with six basic expressions \cite{valstar2010induced}. In the experiments, this dataset was used to validate whether an approximated face manifold can express all the non-linear variations that we want.

\noindent \textbf{LFW dataset:} This dataset consists of 13,233 face images with unconstraint environments such as pose, illumination, expression variations, and occlusion\cite{huang2007labeled}. In the experiments, we used LFW dataset to verify that our generative model could work with unconstrained face images.

\noindent \textbf{MultiPIE dataset:} This dataset consists of 337 subjects with five expressions (e.g., \emph{smile}, \emph{surprise}, \emph{squint}, \emph{disgust}, and \emph{scream}) \cite{gross2010multi}. The expression labels are different from the six basic expressions. This dataset contains face images under pose, illumination and expression variations. In the experiments, we used five head-poses ( $0^{\circ}$, $\pm15^{\circ}$, $\pm30^{\circ}$) and four illumination variations (\emph{dark}, \emph{bright}, \emph{left} and \emph{right} illumination) for head-pose and illumination synthesis.

\begin{figure*}[!ht]
	\begin{center}
		\includegraphics[width=1\linewidth] {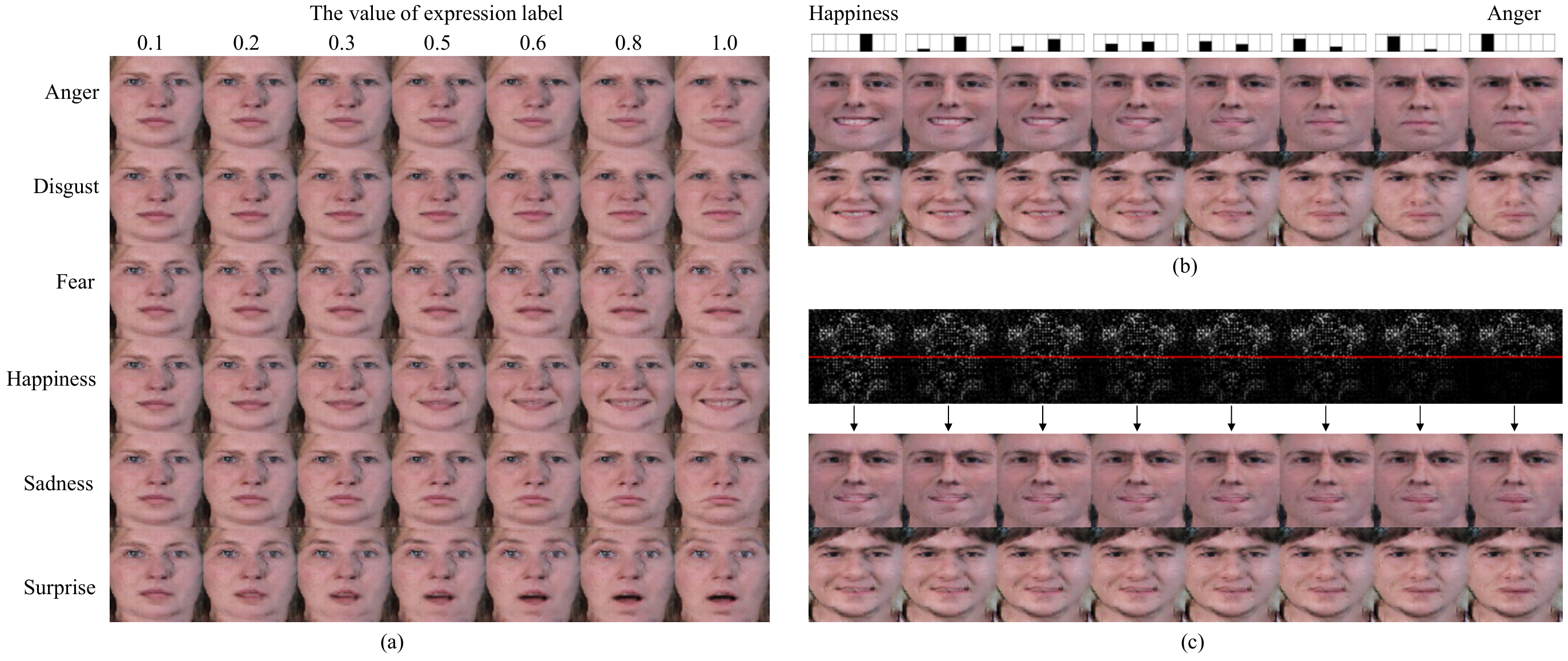}
	\end{center}
	\vspace{-0.45cm}
	\caption{Generated facial expression images by changing the value of target expression label in the label code (a), by changing target expression label from one expression (e.g., happiness) to another expression (e.g., anger) (b), by modifying the label channel (c). In (c), the upper half of the label channel is from anger expression and the bottom half is from happiness. The value of the happiness expression label of bottom half is changed from 1.0 to 0.1. Note that the proposed method can partially control the facial part.}
	\vspace{-0.4cm}
	\label{fig5}
\end{figure*}
\subsection{Experimental settings}
In experiments to verify that our generative model can approximate the face manifold well, we did not perform any traditional data augmentation methods such as linear transformation. Adam optimizer \cite{kingma2014adam} was used with a learning rate of 0.0002 and momentum 0.5. The hyper-parameters $\lambda_\text{diff}$, $\lambda_\text{standard}$, and $\lambda_\text{Recon}$ were empirically selected 0.5, 1.0 and 100, respectively.

\subsection{Qualitative experiments}
\label{sec4.3}In this session, we provide qualitative results to validate our method. We used the PNAS dataset as training dataset, which is a small size dataset (1,510 images). We verified our generative model was effective in approximating the face manifold even with a small size of training data. In experiments, the comparison between proposed method and existing methods was performed with same training dataset (i.e., the PNAS dataset). We used the MMI dataset to validate our generative model.

\subsubsection{The effectiveness of differential discriminator}
\indent

When training data is insufficient, it would be difficult to express various face images through the face manifold and we could not obtain a proper synthesized image. To demonstrate the effectiveness of the differential discriminator, comparative experiments have been conducted.\\
\indent Figure \ref{eq:4} shows facial expression images, which were generated to have target facial expression on facial identity of input image. As shown in Figure \ref{fig4} (a), when training the generator using only standard discriminator, the expression synthesis by traversing face manifold fails. Although the identity is well preserved on the generated image, the target expression is not applied. Figure \ref{fig4} (b) shows the generated images from the generator trained with only differential discriminator. As shown in the figure, the identity is not be maintained while the target expression is synthesized. Since the generator is forced to concentrate only on the facial change between the input and generated image in Eq. \ref{eq:4} and Eq. \ref{eq:5}, the identity could be missed in the face manifold. Finally, Figure \ref{fig4} (c) shows the generated images from the generator trained with both standard discriminator and differential discriminator. It is demonstrated that the face manifold with expression as well as identity of input image is effectively approximated with both discriminators even with a small training dataset. When both discriminators were used together in the experiment, the generator tried to generate a photo-realistic image to fool the standard discriminator, and at the same time, it generated the facial change to fool the differential discriminator.
\begin{figure*}[!ht]
	\begin{center}
		\includegraphics[width=1\linewidth] {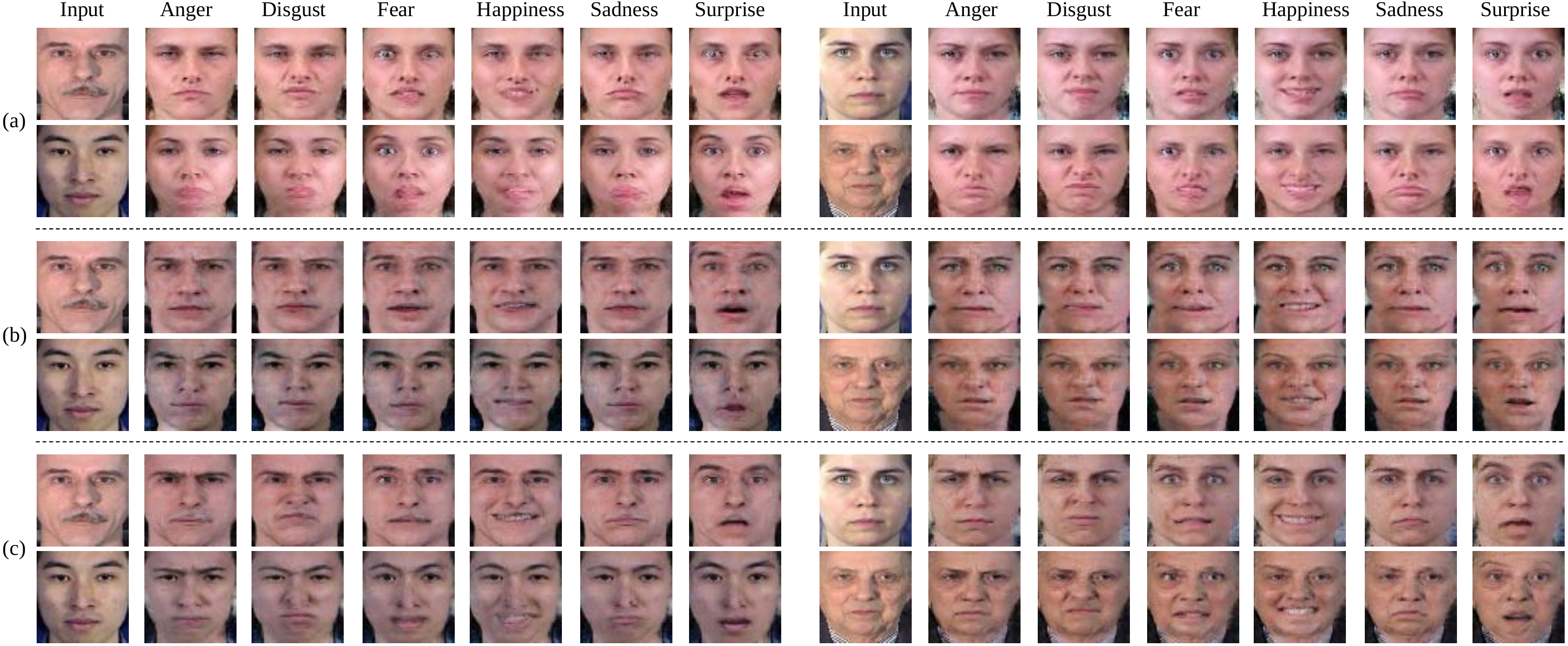}
	\end{center}
	\vspace{-0.45cm}
	\caption{Comparison of generated facial expression images with the methods of CAAE \cite{zhang2017age} (a), Pix2pix \cite{isola2016image} (b), and our proposed model (c). Note that the number of training data was 1,610.}
	\vspace{-0.5cm}
	\label{fig6}
\end{figure*}

\subsubsection{Approximated face manifold}
\indent

We further performed experiments to evaluate that the face manifold was properly approximated for the expression labels that did not exist in the training dataset (e.g., slightly smiled expression, angrily happiness). Figure \ref{fig5} shows the experimental results of approximated face manifold.\\
\indent Figure \ref{fig5} (a) shows the smoothly changed expression images which are generated by changing the value of expression label from 0.1 to 1.0. As shown in the figure, expression intensity of generated images could be controlled by changing the amplitude of expression label.\\
\indent Figure \ref{fig5} (b) shows generated images when our generative model has two expression labels simultaneously. From the left to the right in Figure \ref{fig5} (b), the value of happiness label was set from 1.0 to 0.0 and the value of anger label is set from 0.0 to 1.0. As shown in the figure, the compound expression (e.g., angrily happiness) is generated by using multiple expressions label.\\
\indent In the experiment, we controlled the facial parts such as eyes and lips separately so that useful compound expression could be generated. Figure \ref{fig5} (c) shows generated facial expression images by modifying the label channel. In Figure \ref{fig5} (c), the label channel is made by combining the two label channels of anger and happiness. The upper half of the label channel (above part of the red line) is anger and the lower half of the label channel is from happiness label channel. In the experiment, we gradually reduced the intensity of the lower half of the label channel (happiness label channel). As experimental result, the lip movement of face was changed as shown in the figure. Supplementary video for Figure \ref{fig5} can be viewed at \url{https://youtu.be/GOSuaFpsMTI}.
\vspace{-0.4cm}
\begin{figure}[!ht]
	\centering
	\setlength{\abovecaptionskip}{3pt}
	\setlength{\belowcaptionskip}{-12pt}
	\includegraphics[width=1.0\hsize]{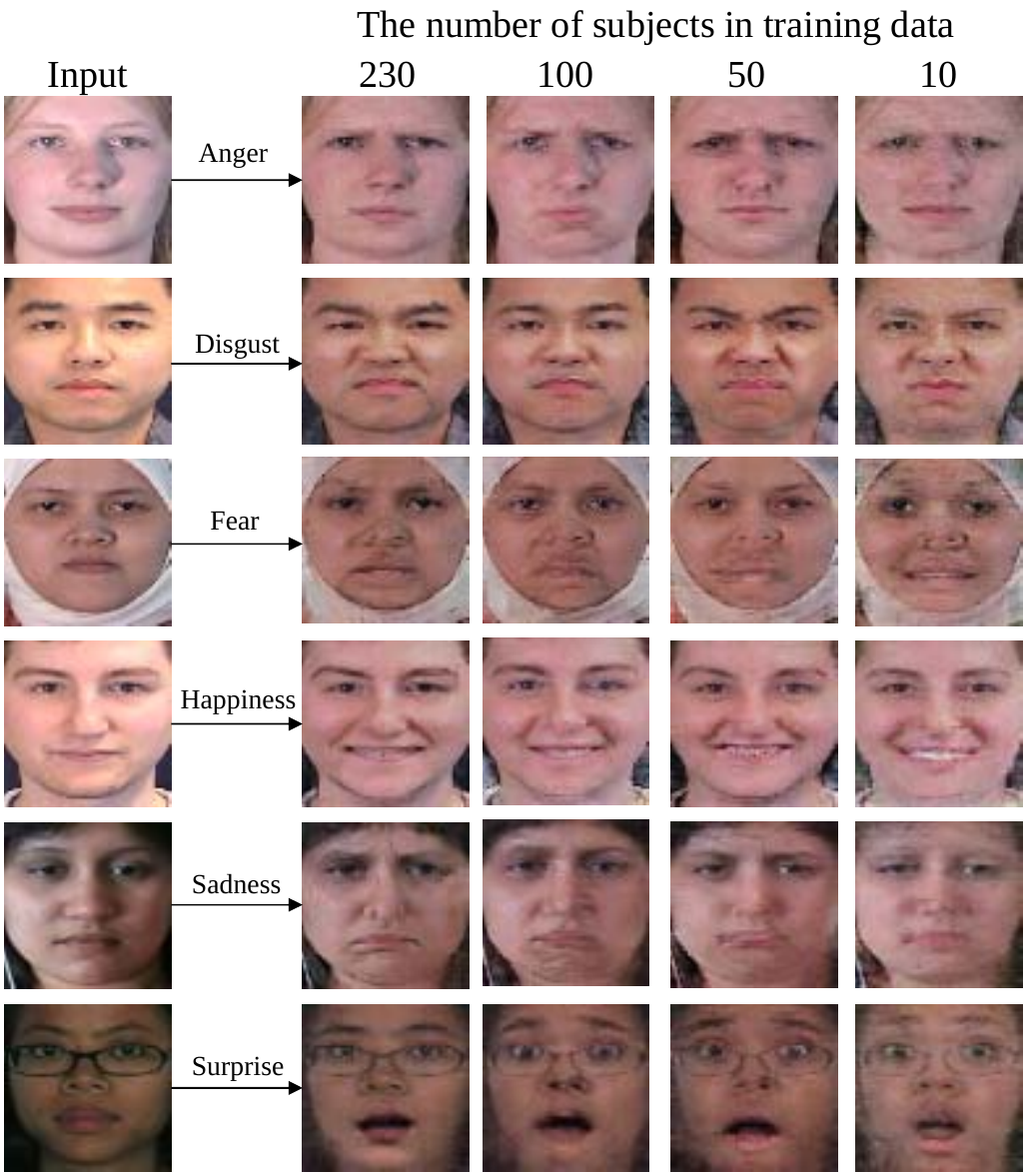}
	\caption{Generated facial expression images with different amounts of training data. Note that input images are not in the training data and expression images are generated while maintaining the identity of the input despite small number of training data.}\label{fig7}
	\vspace{-0.4cm}
\end{figure}
\subsubsection{Comparison with existing models}
\indent

In this session, we compared our generative model with two previous models \cite{zhang2017age,isola2016image}.  CAAE \cite{zhang2017age} was proposed for face aging progression and regression. The authors in \cite{zhang2017age} used 10,670 face images for training CAAE to approximate a face manifold. We modified the number of class labels of CAAE from nine (age labels) to seven (six expressions and neutral expression) for facial expression synthesis. Pix2pix \cite{isola2016image} was reported as a generic model for image-to-image translation. Pix2pix was originally designed for one-directional image translation (e.g., from aerial photos to maps). In order to follow our experimental scenario (multi-directional image translation), we slightly modified the structure of Pix2pix by adding label channel. As shown in Figure \ref{fig6}, our generative model preserved the identity of input image and synthesized target facial expression well even with small size (1,610) of face images.
\begin{figure*}[!ht]
	\begin{center}
		\includegraphics[width=1\linewidth] {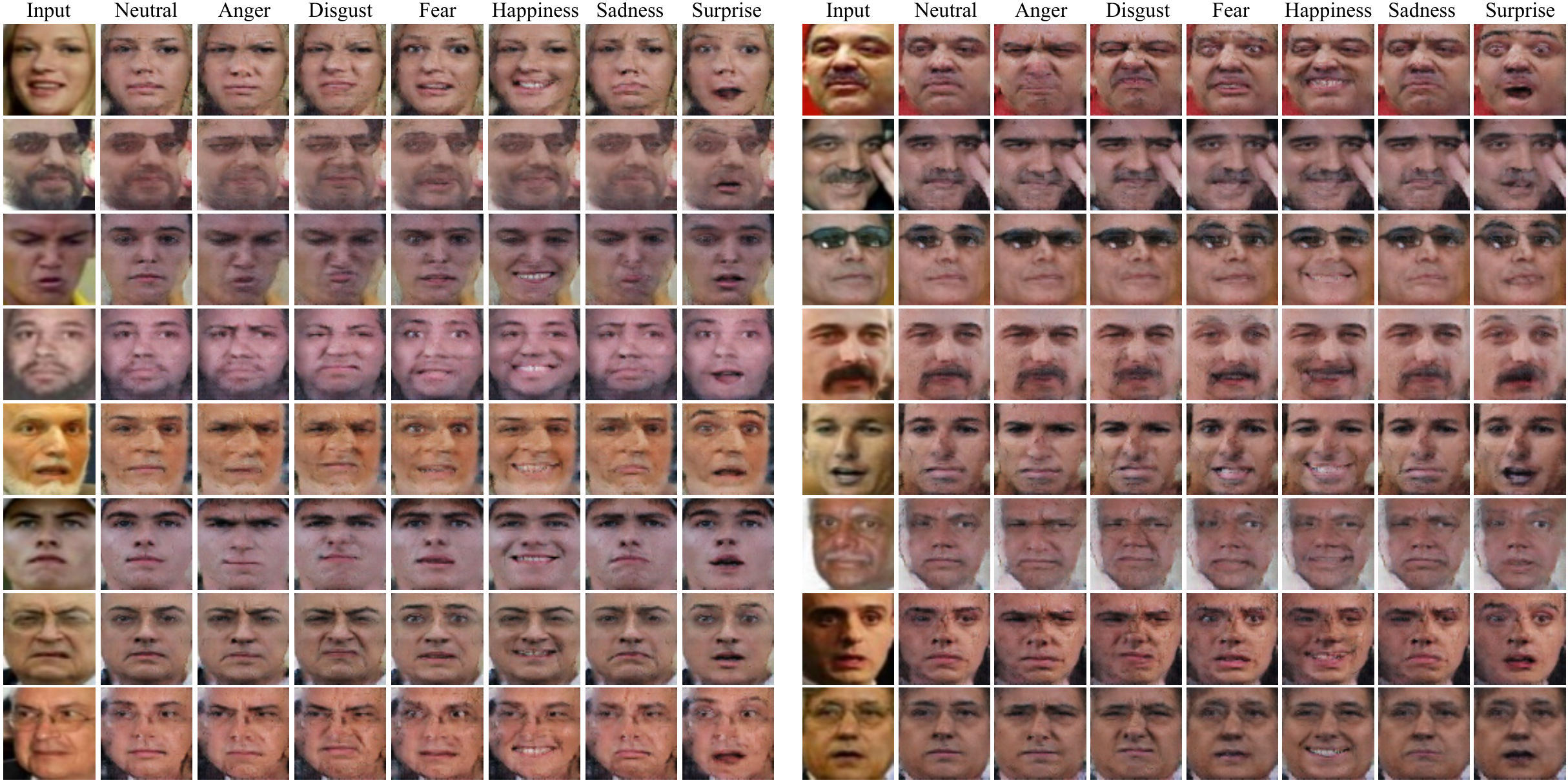}
	\end{center}
	\vspace{-0.45cm}
	\caption{Generated facial expression images from a wild test input of the LFW dataset. Note that our generator is trained with frontal face images.}
	\vspace{-0.4cm}
	\label{fig8}
\end{figure*}
\begin{figure*}[!ht]
	\begin{center}
		\includegraphics[width=1\linewidth] {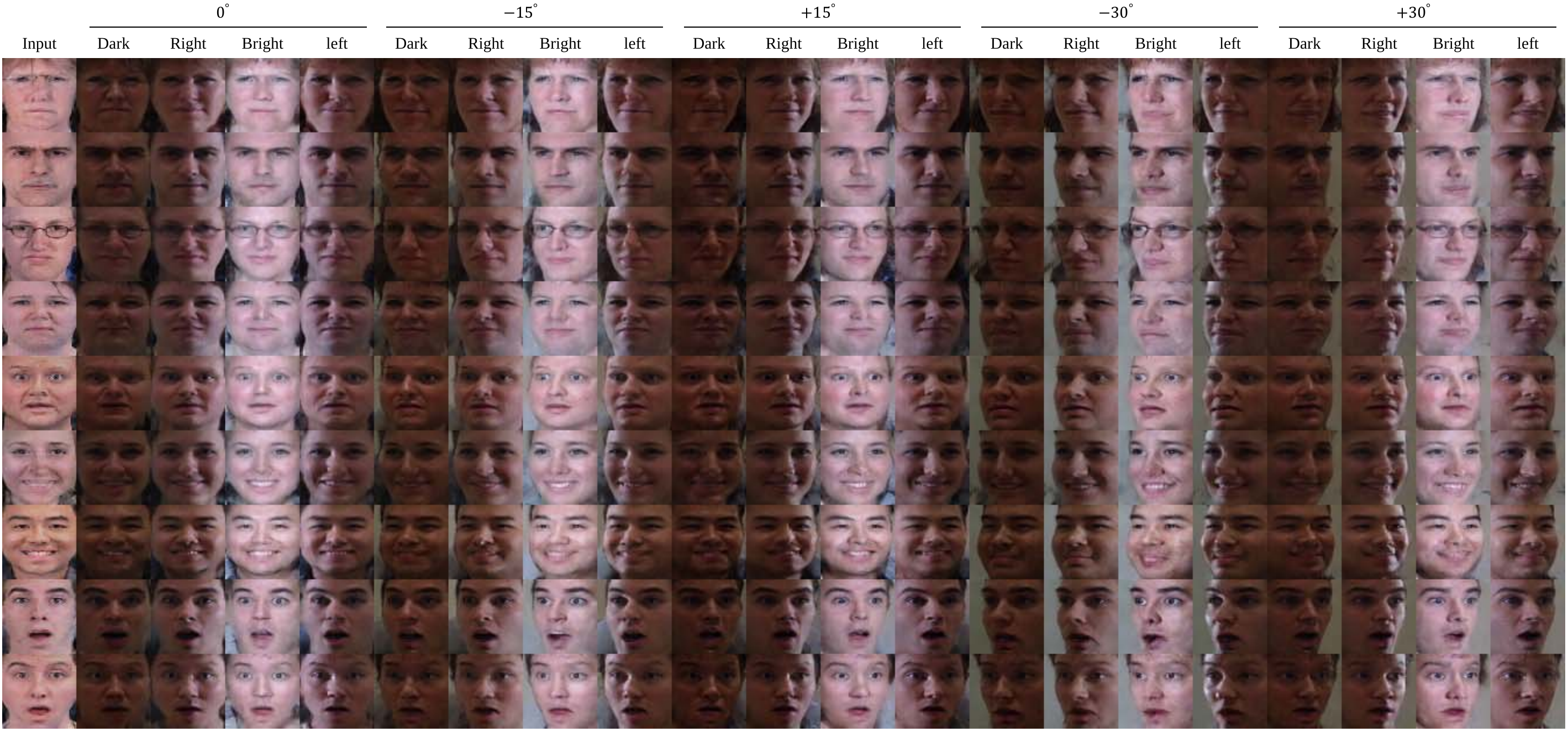}
	\end{center}
	\vspace{-0.45cm}
	\caption{Generated facial expression images with synthesized head-pose and illumination variations by the proposed method.}
	\vspace{-0.5cm}
	\label{fig9}
\end{figure*}

\subsubsection{Experimental results with different amount of training data}
\indent

In order to demonstrate the effectiveness of the proposed method under small amount of training data, we further performed experiments with different amount of training data. Four different amounts of training data were used.  One is the PNAS dataset, which consists of 230 (1610 images) subjects. The other three dataset respectively consist of 100 (700 images), 50 (350 images), and 10 (70 images) subjects, which were randomly sampled from the PNAS dataset. For fair comparison, all the networks trained with four datasets had same structure and trained from scratch without any sophisticated training strategy \cite{larochelle2009exploring,pan2010survey,ding2017exprgan,shrivastava2016learning}. In this experiment, we did not use any data augmentation in training dataset.\\
\indent As shown in Figure \ref{fig7}, an artifact with appearance change is increasing as the number of subjects is decreasing. Nevertheless, our methods effectively approximate the face manifold, even with the very small amount of dataset (10 subjects with 70 images). The generated images for the entire subjects in the MMI dataset are shown in the Appendix \ref{appendix}.

\subsubsection{Experimental results on LFW dataset}
\indent

LFW dataset \cite{huang2007labeled} was designed for face verification with unconstrained environments such as pose, illumination, expression variations, and occlusion. As we used LFW dataset as an input of the generator, we evaluated that our model could synthesize the facial expression on wild face images. Figure \ref{fig8} shows the generated facial expression images, which reasonably maintain pose and occlusion as well as identity of input image. In case of \emph{over-posed} image or \emph{irregular} distribution of illumination image, artifacts could be seen. These artifacts could be raised because the training data (PNAS dataset) was collected under well-distributed and frontal face image.

\subsubsection{Experiment for synthesizing head-pose and illumination variations}
\indent

In order to validate the generality of our proposed method, we performed experiments to synthesize head-pose and illumination variations as well as facial expression synthesis. We used five head poses ( $0^{\circ}$, $\pm15^{\circ}$, $\pm30^{\circ}$), and four illumination variations (dark, bright, left, and right illuminations) in MultiPIE dataset. We randomly sampled 50 subjects out of 337 subjects in MultiPIE dataset and the number of the sampled dataset was 2,000. We slightly changed our network architecture to receive two labels (i.e., pose label and illumination label): Two label codes, which refer to head-pose and illumination label, were employed to make a label channel. Except that, all the structure of network and the hyper-parameters were same with the expression synthesis experiments described above. The generated facial expression with subjects of MMI dataset were used as inputs to subsequently synthesize head-pose and illumination variations. Figure \ref{fig9} shows the generated expression images with the synthesized head-pose and illumination variations, which maintain appearance of input image.

\subsection{Quantitative experiments}
In this session, we verified the usefulness of the generated images by the proposed method. We quantitatively evaluated the performance of a facial expression classifier by augmenting training data with D-GAN. The non-linear data augmentation by the proposed method was compared with the linear augmentation in terms of accuracy for facial expression recognition. We took a simple deep neutral networks (i.e., AlexNet \cite{krizhevsky2012imagenet}) for the facial expression classifier and MMI dataset to validate the classifier. The MMI dataset included individuals who posed expressions non-uniformly, wore glasses or caps, and had mustaches or head movements. Therefore, the facial expression recognition task on MMI dataset was relatively challenging \cite{mollahosseini2016going,kim2017deep,liu2014learning}. The MMI dataset consists of 312 sequences with six basic expressions. Most subjects do not have all six basic expressions and neutral expression. The sequences start with the neutral expression, passes through the peak expression, and return to the neutral expression. We used 205 sequences captured in a front view. We extracted 5 or 7 neutral expression images per subject and 7 peak expression images per sequence. The total number of extracted MMI dataset was 1,689. We divided the MMI dataset into 10 subject-independent folds. The accuracy of facial expression recognition was reported by 10-fold cross validation. As described in Table \ref{table1}, the accuracy of facial expression recognition without any augmentation method was $55.89\%$.\\
\indent We used linear augmentation methods such as rotation ( $\pm3^{\circ}$, $\pm5^{\circ}$) and translation ( $\pm2$, $\pm4$ ), and total 28 images per one image were augmented. We augmented 1,689 images to 48,981 images using linear augmentation methods. The accuracy of facial expression recognition with linearly augmented data was $62.40\%$.\\
\indent We generated 210 facial expression images using the subjects of MMI dataset to make all the subjects in MMI dataset have basic expressions, and 40,796 images using the subjects of LFW dataset to augment various subjects, head-pose, and occlusions. The total number of MMI dataset with generated images was 42,698. The result of 10-fold cross validation on non-linearly augmented data was $68.20\%$. Note that the generated images of subjects in test fold were not used for training and testing the facial expression classifier. Experimental results showed that non-linear augmentation (with D-GAN) in training data could achieve the performance improvement of $5.8\%$ compared with linear augmentation method.

\input{./tables/table1.tex}

\section{Conclusion}
In this paper, we proposed the differential generative adversarial networks (D-GAN) to approximate the face manifold for non-linear augmentation, even in a small amount of dataset. The differential discriminator, which complemented the standard discriminator, was devised to make the generator focus on the facial change between input and output image. Through comprehensive experiments, we demonstrated our proposed method could effectively approximate the face manifold. Moreover, we showed that our proposed method could synthesize the facial image for the combined target label, which was unseen in training data. Through quantitative experiment, we demonstrated that non-linear augmented data through our proposed method could improve the facial expression classifier.

{\small
\bibliographystyle{ieee}
\bibliography{egbib}
}



\clearpage
\onecolumn
\section{Appendix}
\label{appendix}

\begin{figure*}[ht]
	\small
	\begin{center}
		\includegraphics[width=1.0\linewidth] {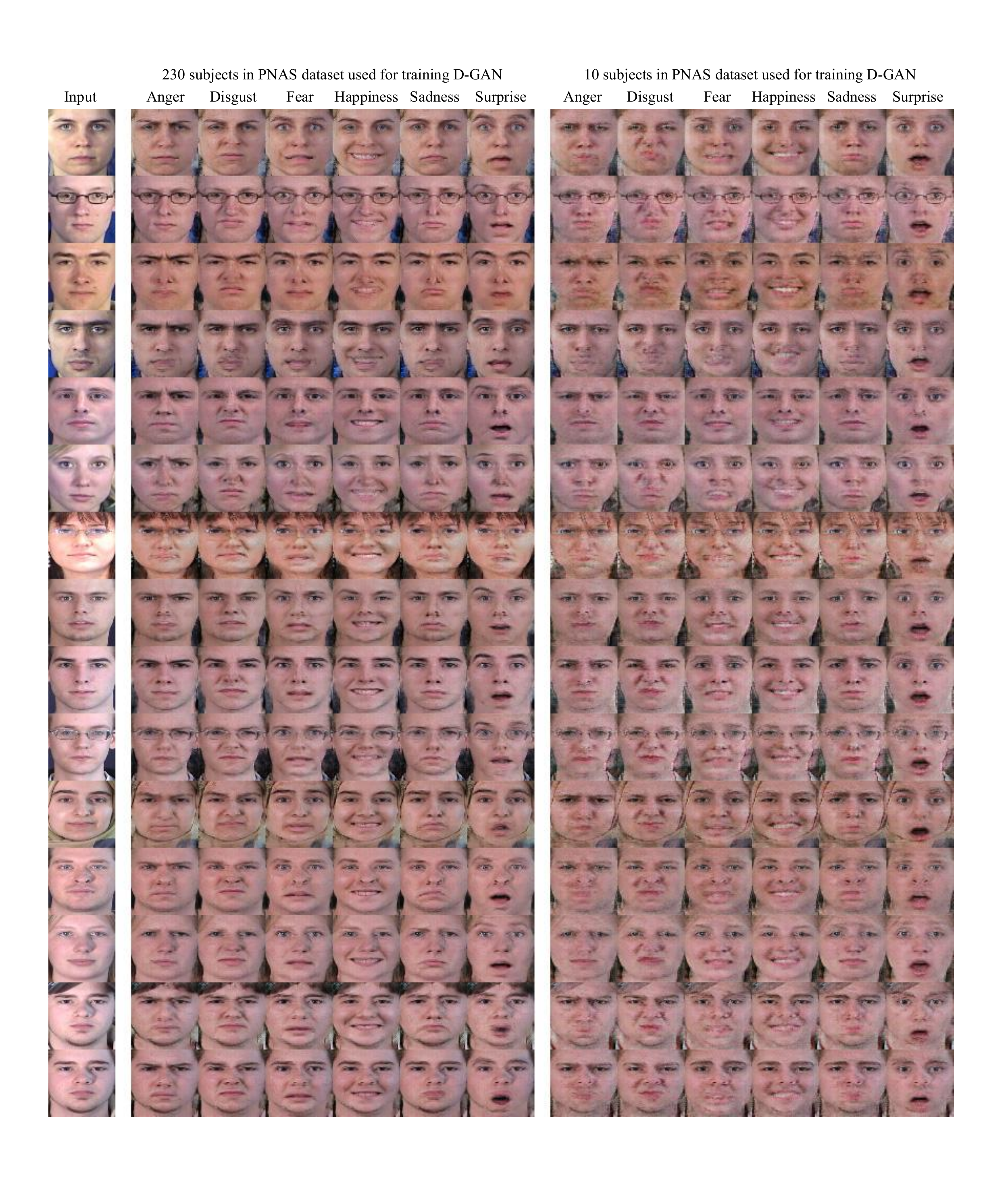}
		\vspace{-0.8cm}
	\end{center}
	\caption{Additional generated images using MMI dataset as input. The first column is the input image. The second through seventh columns are the result of training using all PNAS datasets (1,610 images). The eighth through thirteenth colums are the result of training using only 10 subjects (70 images) in the PNAS dataset.}
	\vspace{-0.2cm}
	\label{supp_fig1}
\end{figure*}

\begin{figure*}[!ht]
	\small
	\begin{center}
		\includegraphics[width=1.0\linewidth] {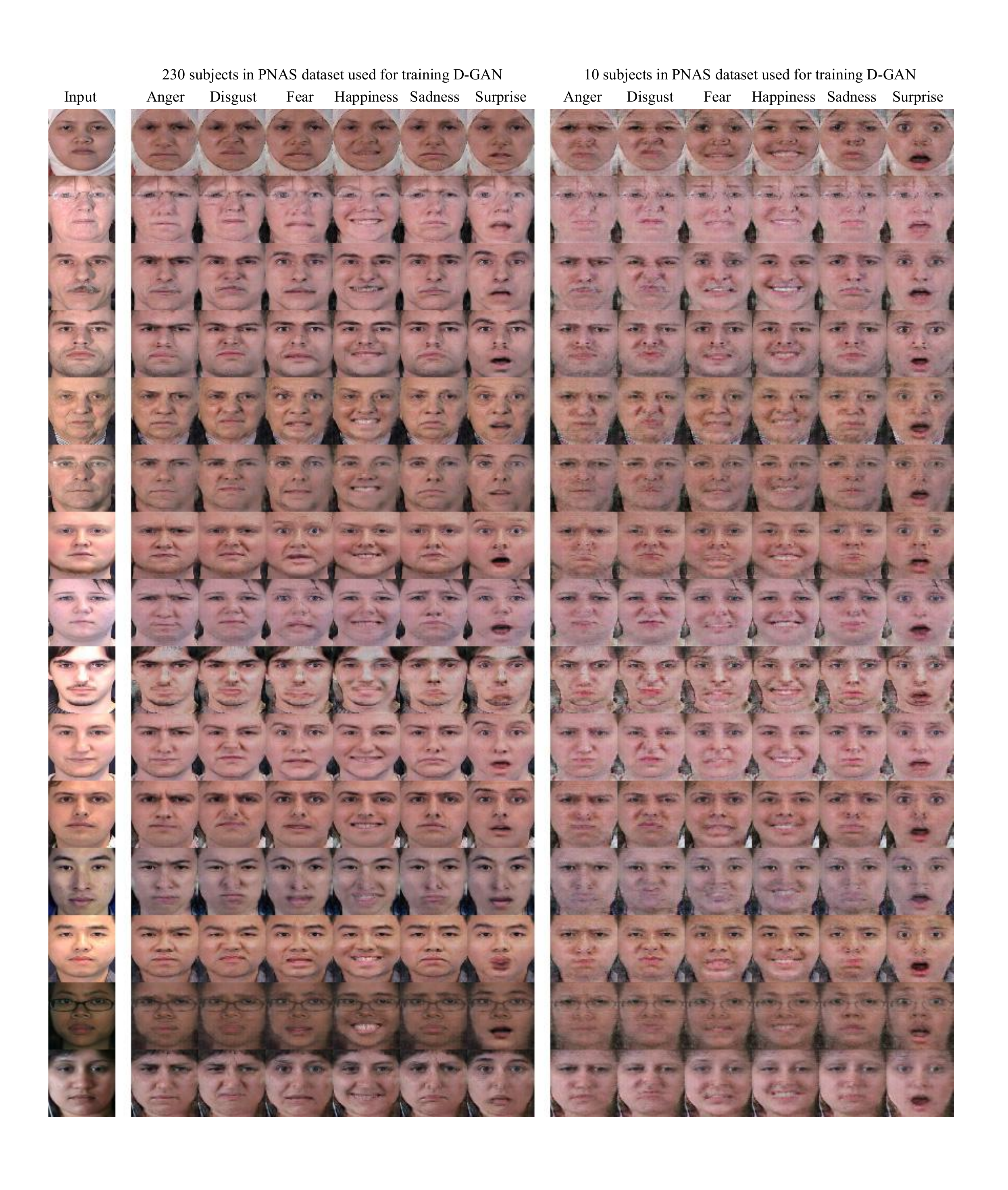}
		\vspace{-0.8cm}
	\end{center}
	\caption{Additional generated images using MMI dataset as input. The first column is the input image. The second through seventh columns are the result of training using all PNAS datasets (1,610 images). The eighth through thirteenth columns are the result of training using only 10 subjects (70 images) in the PNAS dataset.}
	\vspace{-0.2cm}
	\label{supp_fig2}
\end{figure*}

\begin{figure*}[!ht]
	\small
	\begin{center}
		\vspace{-1.6cm}
		\includegraphics[width=1.0\linewidth] {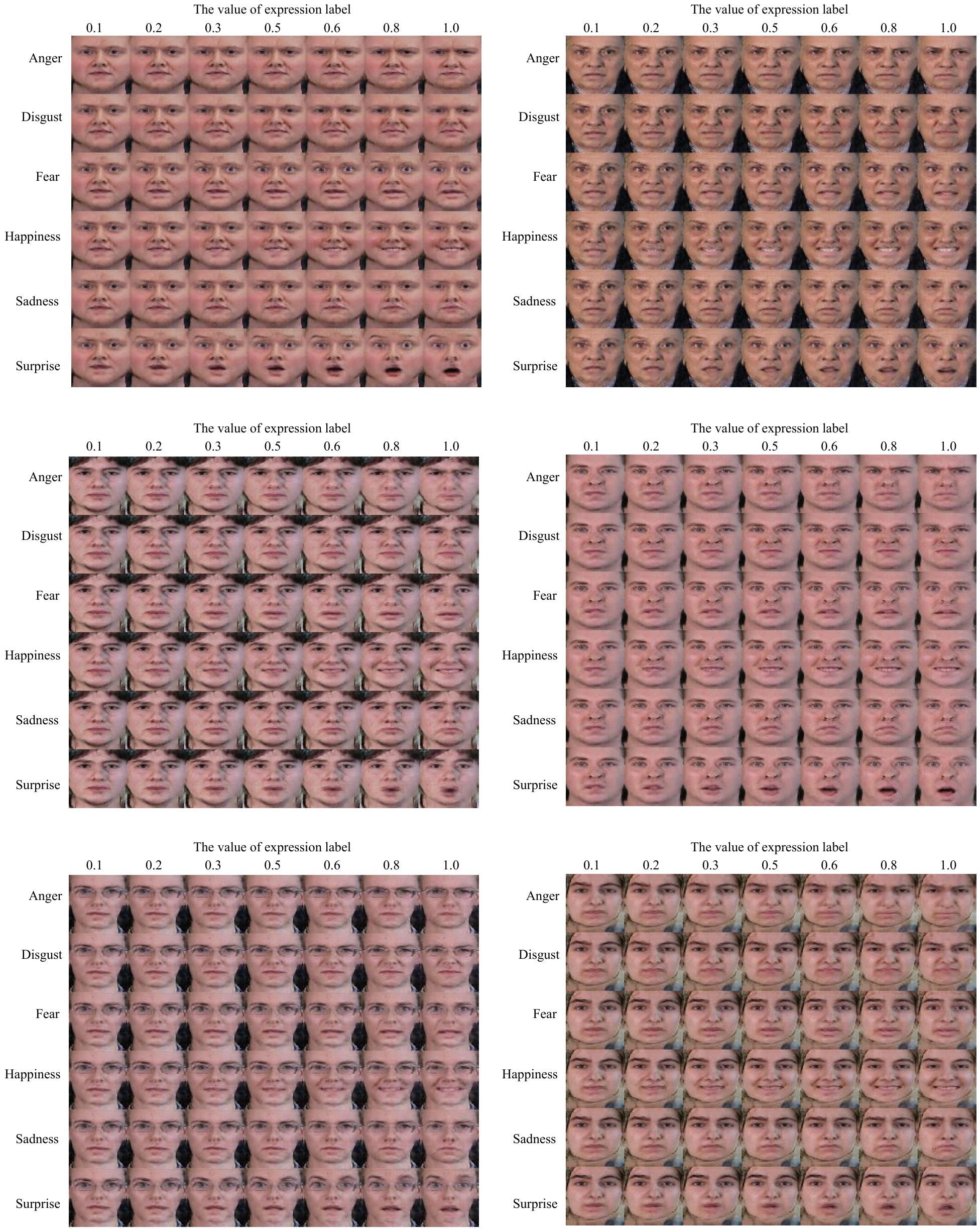}
		\vspace{-0.8cm}
	\end{center}
	\caption{Additional generated images by changing the value of target expression label in the label code.}
	\vspace{-0.2cm}
	\label{supp_fig3}
\end{figure*}
\begin{figure*}[!ht]
	\small
	\begin{center}
		\vspace{-1.6cm}
		\includegraphics[width=1.0\linewidth] {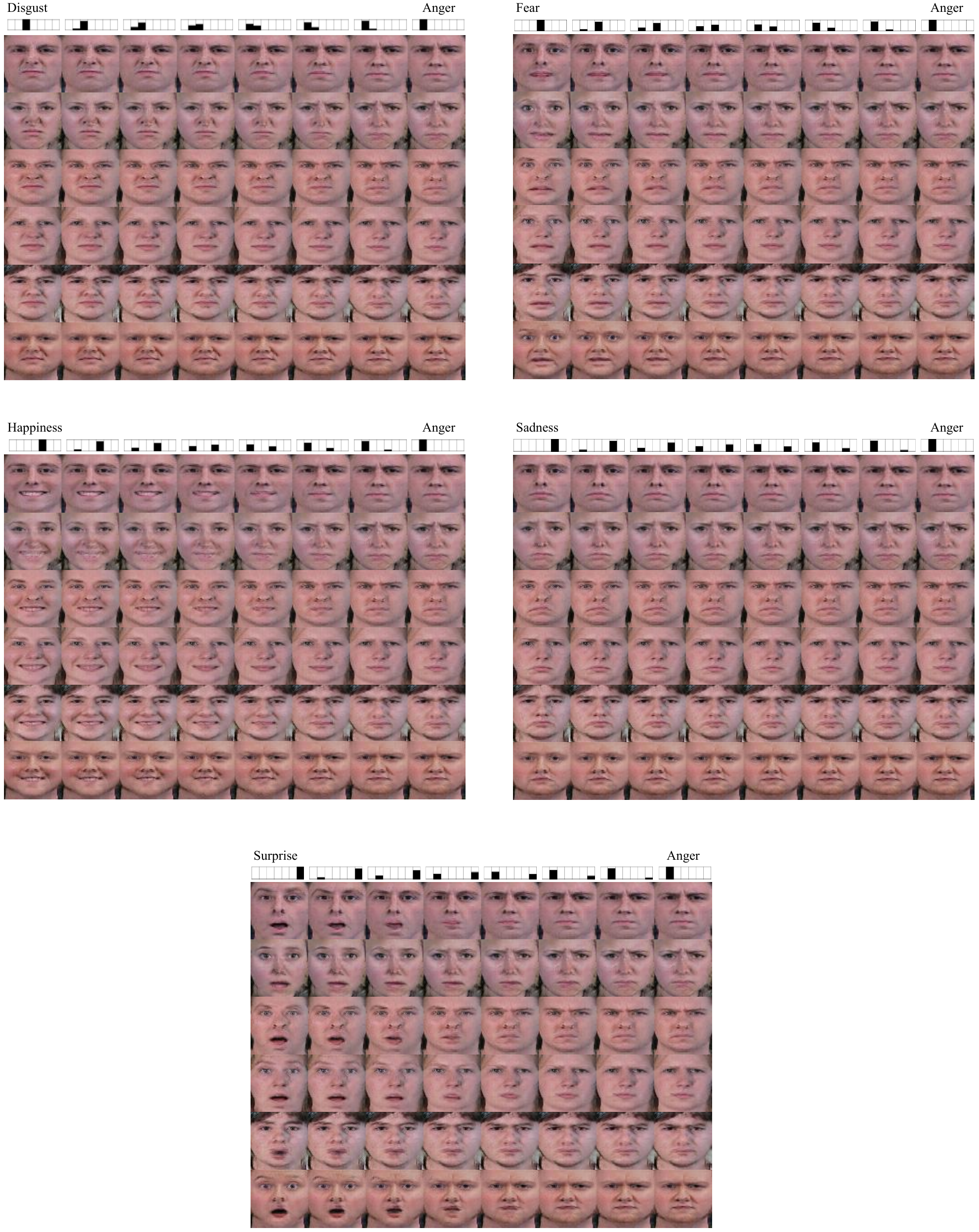}
		\vspace{-0.8cm}
	\end{center}
	\caption{Additional generated images by changing the value of target expression label in the label code from one expression to another expression.}
	\vspace{-0.2cm}
	\label{supp_fig4}
\end{figure*}
\begin{figure*}[!ht]
	\small
	\begin{center}
		\vspace{-1.6cm}
		\includegraphics[width=1.0\linewidth] {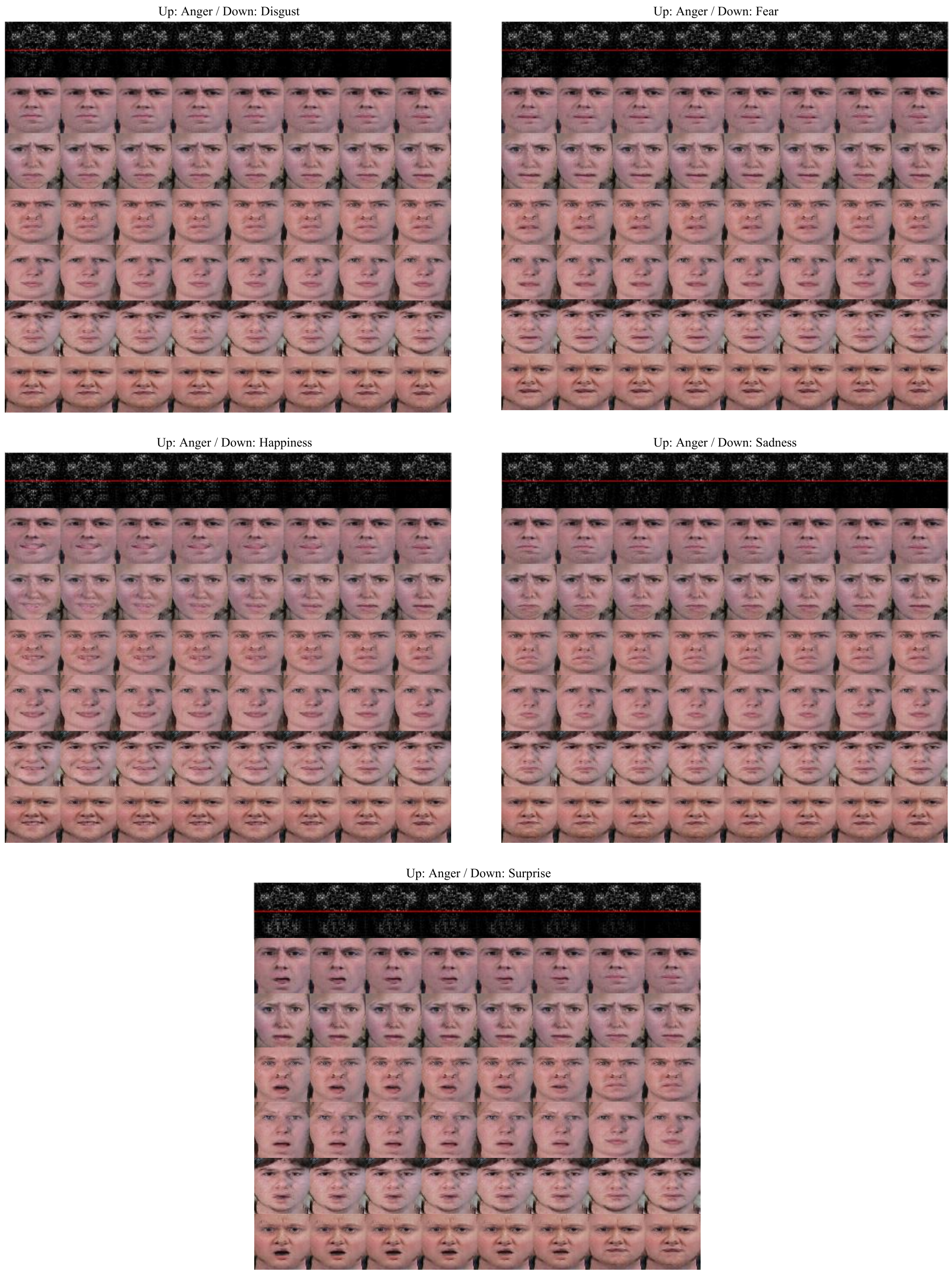}
		\vspace{-0.8cm}
	\end{center}
	\caption{Additional generated images by modifying the label channel. The upper half of the label channel is from anger expression and we changed the bottom half from other expression. The value of the other expression
		label of the bottom half is changed from 1.0 to 0.1.}
	\vspace{-0.2cm}
	\label{supp_fig5}
\end{figure*}
\begin{figure*}[!ht]
	\small
	\begin{center}
		\vspace{-1.6cm}
		\includegraphics[width=1.0\linewidth] {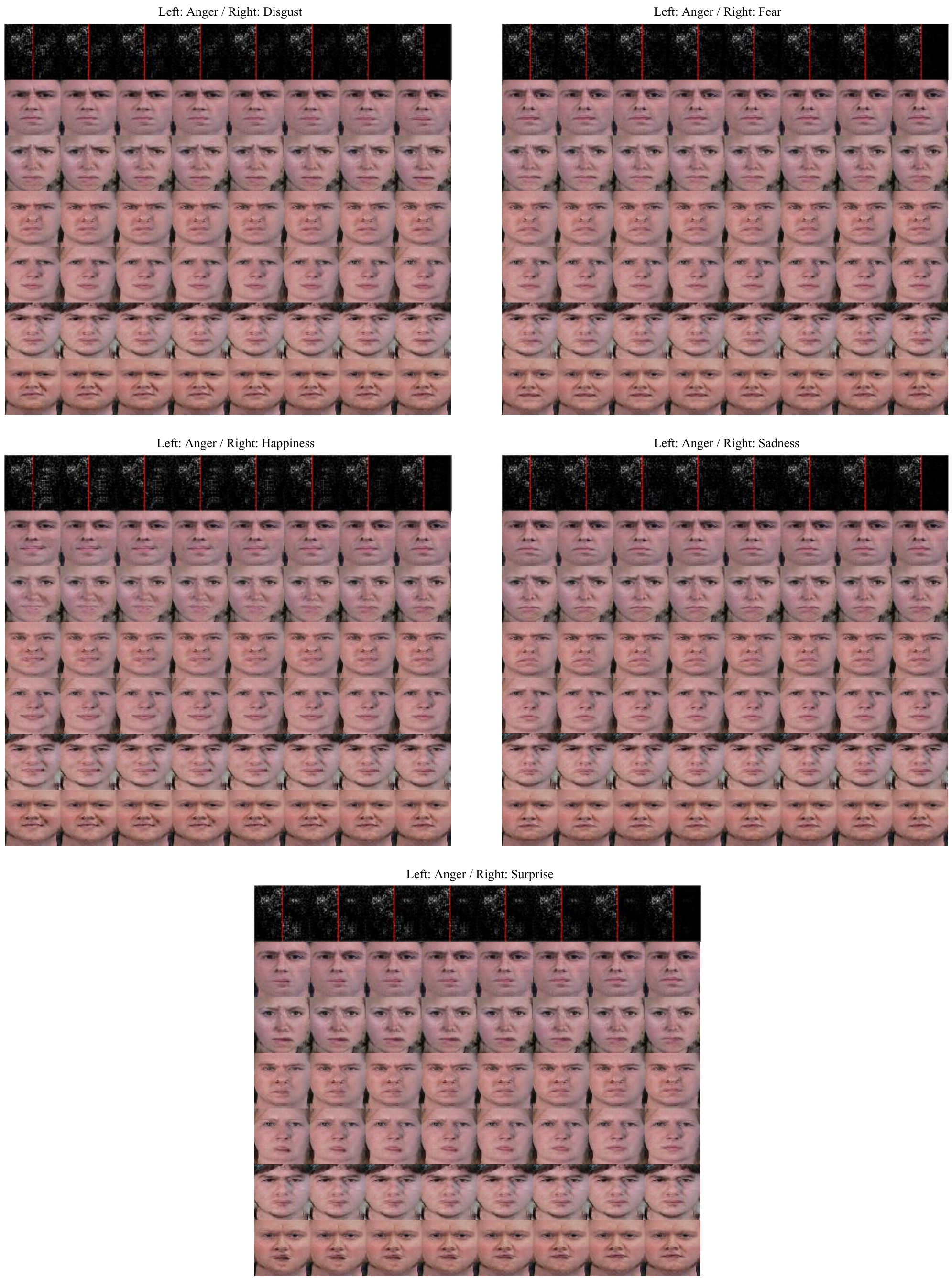}
		\vspace{-0.8cm}
	\end{center}
	\caption{Additional generated images by modifying the label channel. The left half of the label channel is from anger expression and we changed the right half from other expression. The value of the other expression
		label of the right half is changed from 1.0 to 0.1}
	\vspace{-0.2cm}
	\label{supp_fig6}
\end{figure*}

\begin{figure*}[!ht]
	\small
	\begin{center}
		\includegraphics[width=1.0\linewidth] {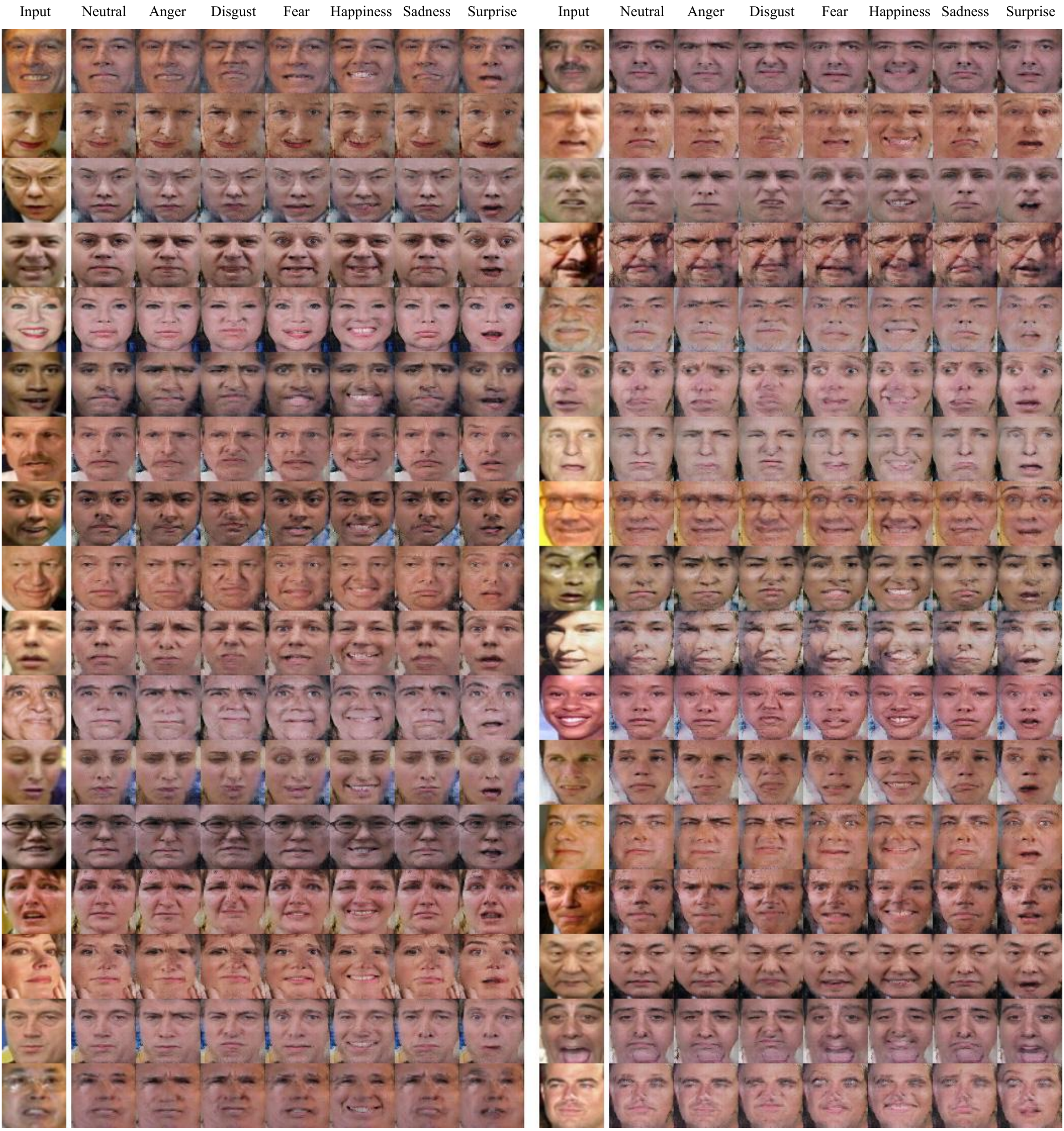}
		\vspace{-0.8cm}
	\end{center}
	\caption{Additional generated images using LFW dataset as input. Note that our generator is trained with frontal face images (PNAS dataset)}
	\vspace{-0.2cm}
	\label{supp_fig7}
\end{figure*}

\begin{figure*}[!ht]
	\small
	\begin{center}
		\includegraphics[width=1.0\linewidth] {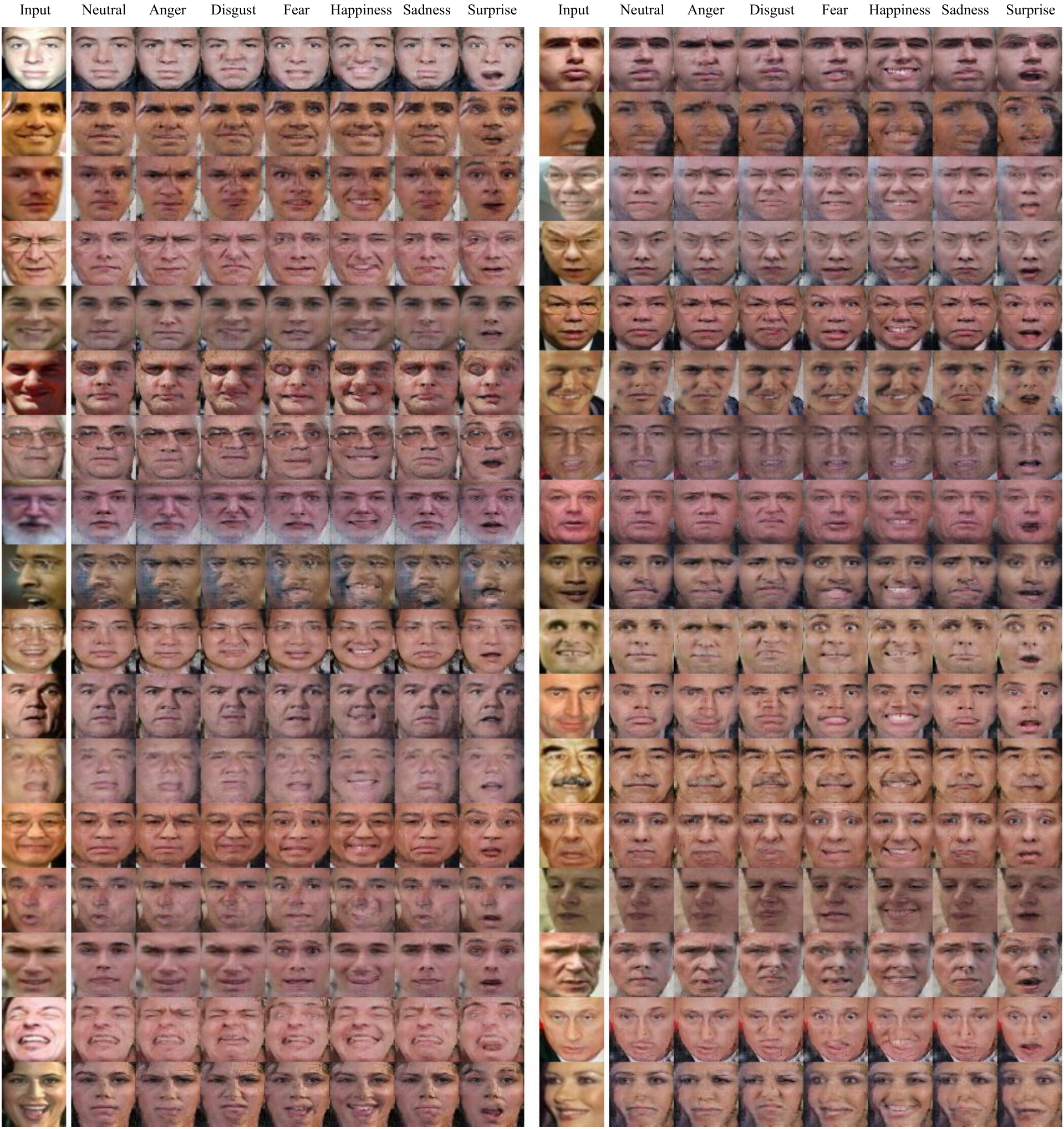}
		\vspace{-0.8cm}
	\end{center}
	\caption{Additional generated images using LFW dataset as input. Note that our generator is trained with frontal face images (PNAS dataset)}
	\vspace{-0.2cm}
	\label{supp_fig8}
\end{figure*}

\begin{figure*}[!ht]
	\small
	\begin{center}
		\includegraphics[width=1.0\linewidth] {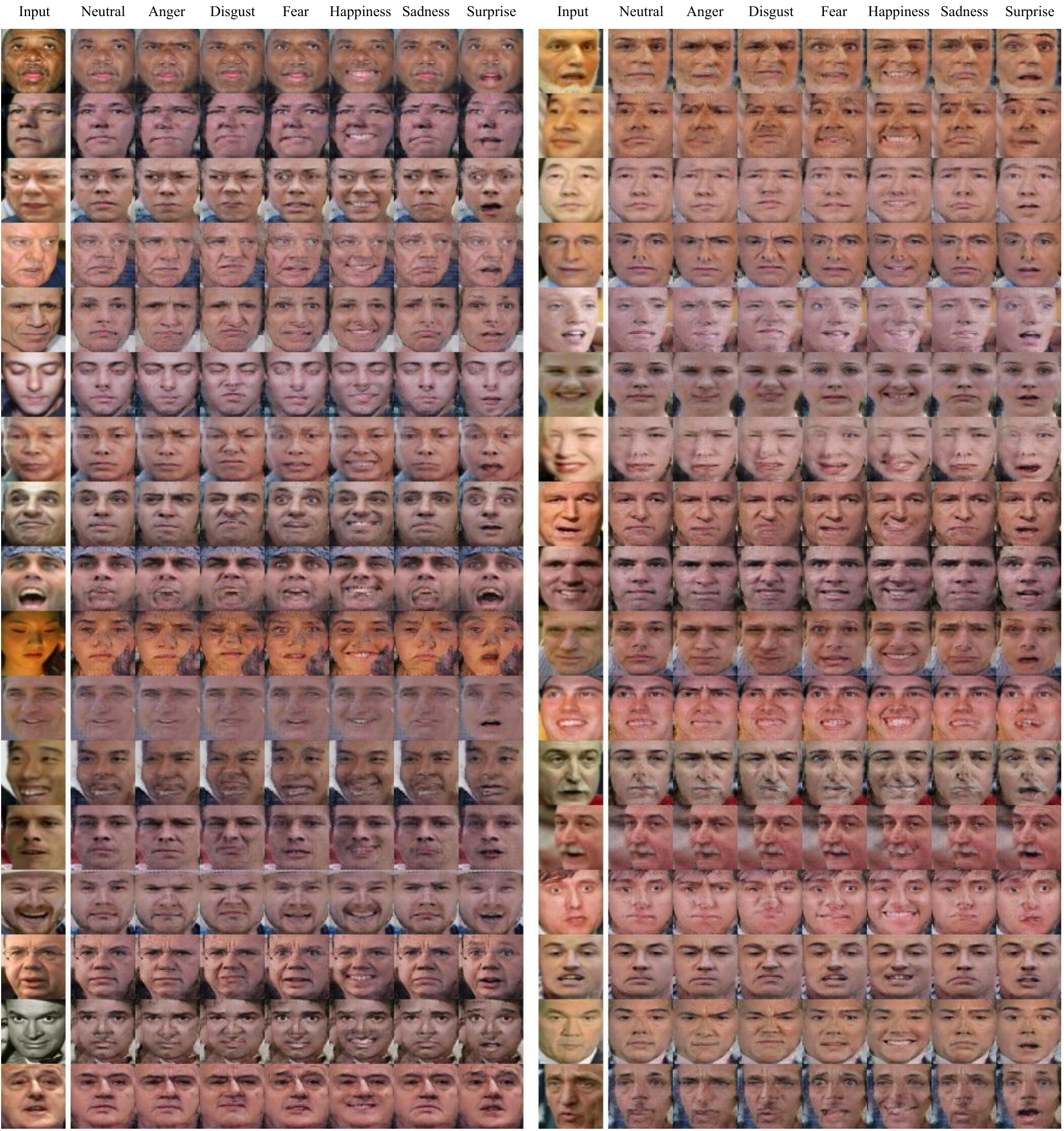}
		\vspace{-0.8cm}
	\end{center}
	\caption{Additional generated images using LFW dataset as input. Note that our generator is trained with frontal face images (PNAS dataset)}
	\vspace{-0.2cm}
	\label{supp_fig9}
\end{figure*}

\begin{figure*}[!ht]
	\small
	\begin{center}
		\includegraphics[width=1.0\linewidth] {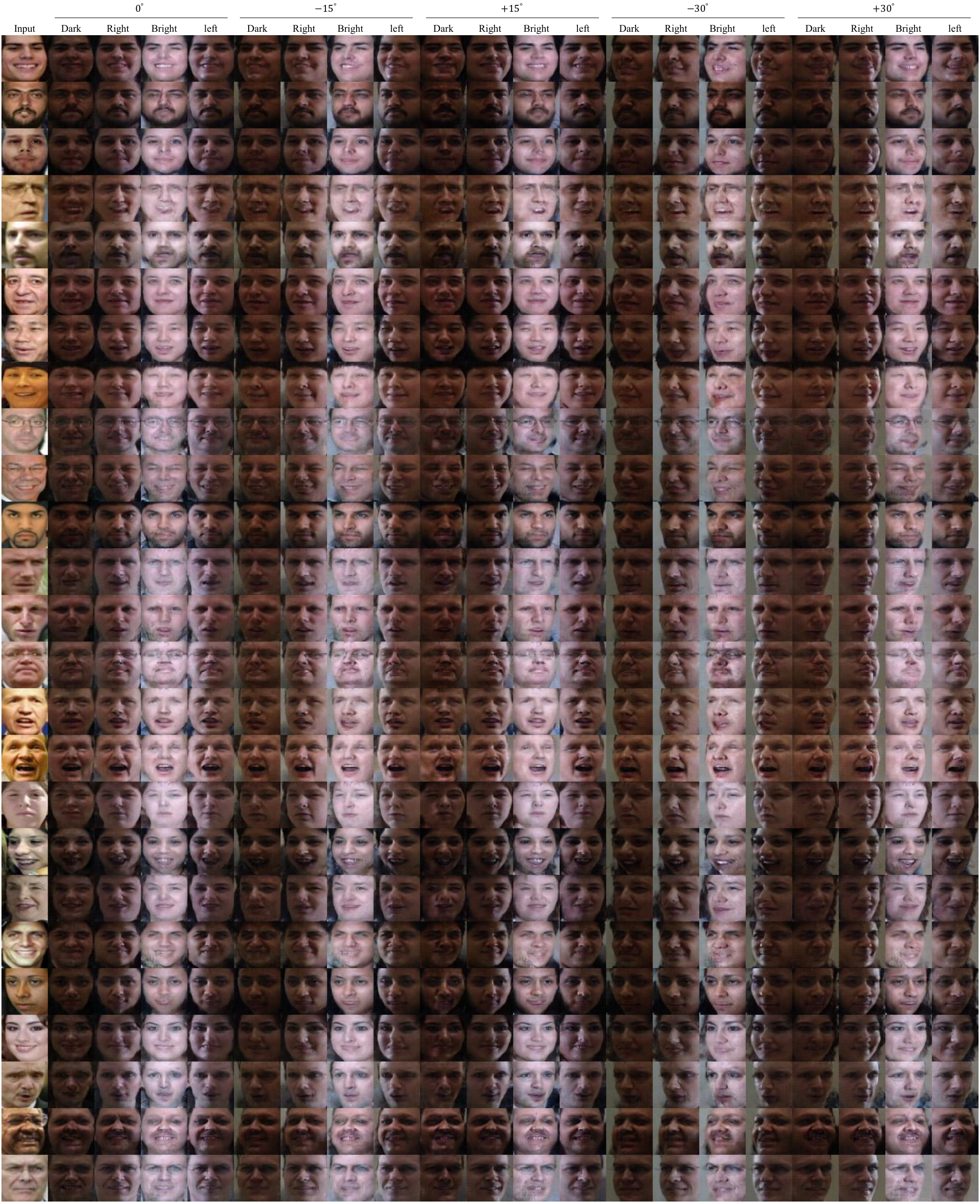}
		\vspace{-0.8cm}
	\end{center}
	\caption{Additional generated images for head-pose and illumination synthesis using LFW dataset as input.}
	\vspace{-0.2cm}
	\label{supp_fig10}
\end{figure*}

\end{document}

%% file: tables/table1.tex
\begin{table}
	\begin{center}
	{\small
		\begin{tabular}{l c c}
			\hline
			\textbf{Method} & \textbf{\# of images} & \textbf{Accuracy(\%)} \\
			\hline
			\hline	
			MMI  &  1,689 & 55.89  \\
			MMI with Linear augmentation  &  48,981 & 62.40 \\
			MMI with Non-linear & &\\
			augmentation by D-GAN  & 42,698 & \textbf{68.20}\\
			\hline	
		\end{tabular}
	}	
	
	\caption{Quantitative results.}
	\label{table1}
	\end{center}
\vspace{-0.65cm}
\end{table}